\documentclass[runningheads]{llncs}
\usepackage[T1]{fontenc}
\usepackage{hyperref}
\usepackage{graphicx}
\usepackage{amsmath}
\usepackage{xcolor}
\usepackage{algorithm}
\usepackage{algorithmic}
\usepackage{pifont}
\usepackage{multicol}
\usepackage{comment}
\usepackage{multirow}
\usepackage{wrapfig,lipsum,booktabs}
\newcommand{\etal}{\textit{et al}.}

\begin{document}
\title{Evaluation of Convolution Primitives for Embedded Neural Networks on 32-bit Microcontrollers}
\titlerunning{Evaluation of Convolution Primitives on ARM-Cortex M}
%
\author{Baptiste Nguyen\inst{1,2} \and
Pierre-Alain Moëllic\inst{1,2} \and
Sylvain Blayac\inst{3}}
\authorrunning{B. Nguyen et al.}
%

\institute{CEA Tech, Centre CMP, Equipe Commune CEA Tech - Mines Saint-Etienne, F-13541 Gardanne, France
\and 
Univ. Grenoble Alpes, CEA, Leti, F-38000 Grenoble, France\\
\email{baptiste.nguyen@cea.fr, pierre-alain.moellic@cea.fr}\\
\and
Mines Saint-Etienne, CMP, Department of Flexible Electronics, F-13541 Gardanne, France\\
\email{blayac@emse.fr}\\
}
\maketitle              
\begin{abstract}
Deploying neural networks on constrained hardware platforms such as 32-bit microcontrollers is a challenging task because of the large memory, computing and energy requirements of their inference process. To tackle these issues, several convolution primitives have been proposed to make the standard convolution more computationally efficient. However, few of these primitives are really implemented for 32-bit microcontrollers. In this work, we collect different state-of-the-art convolutional primitives and propose an implementation for ARM Cortex-M processor family with an open source deployment platform (NNoM). Then, we carry out experimental characterization tests on these implementations. Our benchmark reveals a linear relationship between theoretical MACs and energy consumption. Thus showing the advantages of using computationally efficient primitives like shift convolution. We discuss about the significant reduction in latency and energy consumption due to the use of SIMD instructions and highlight the importance of data reuse in those performance gains. For reproducibility purpose and further experiments, codes and experiments are publicly available\footnote{\url{https://gitlab.emse.fr/b.nguyen/primitive_of_convolution}}.
\end{abstract}
\section{Introduction}
The demand for edge inference is growing and neural networks are prime candidates due to their success across a large variety of application domains. However, state-of-the-art deep neural network models, especially convolution neural networks, require a large amount of memory and computational resources. For example, the standard ResNet-18 model~\cite{he2016deep} for image classification on ImageNet has around 11M parameters and requires approximately 1 GMACs for an inference which is prohibitive for ARM Cortex-M microcontrollers. Thus, designing efficient neural network architectures is a major topic in the embedded AI community. In the search for efficient neural network architectures, several alternatives to convolution have been proposed, but few of them are practically implemented on deployment libraries for 32-bit microcontrollers. This work focuses on the implementation and characterization of state-of-the-art convolution primitives for ARM Cortex-M MCUs.
\textbf{Our contributions are as follow:}
\begin{itemize}
\item We implement three state-of-the-art convolution primitives for ARM Cortex-M MCUs and when possible, we propose another implementation which makes use of the SIMD\footnote{\url{https://www.keil.com/pack/doc/CMSIS/Core/html/group\_\_intrinsic\_\_SIMD\_\_gr.html}} instructions (\textit{Single Instruction, Multiple Data}).
\item We characterize the latency and energy consumption of five primitives, including the standard convolution, against different parameters such as kernel or input size.
\item We provide insights on the performance of different primitives, especially for our implementations using SIMD instructions to help machine learning practitioners to design, develop and deploy efficient models according to their requirements. \end{itemize}

\section{Background}
\subsection{Preliminaries and notation}
We consider the typical case of a 2D-convolution layer with padding and a square input tensor $X$ of dimensions of $H_x\times H_x\times C_x$ with $H_x$ the spatial width and $C_x$ the number of channels. The convolution layer produces an output tensor $Y$ of dimensions $H_y\times H_y\times C_y$ with $H_y$ the spatial width (equal to $H_x$) and $C_y$ the number of channels. The convolution is performed thanks to convolutional kernels represented by a weight tensor $W$ of size $H_k\times H_k\times C_x\times C_y$ with $H_k$ the spatial dimension of a kernel (assumed to be square), $C_x$ the number of input channels and $C_y$ the number of output channels (i.e. the number of filters) as defined previously. The output for standard convolution is as follows:
\begin{equation}
Y_{k,l,n}=\sum_{m=1}^{C_x}\sum_{i=1}^{H_k}\sum_{j=1}^{H_k}W_{i,j,m,n}\cdot X_{k+i-1,l+j-1,m}\quad\forall k,l \in [1,H_y],\quad\forall n \in [1,C_y]
\end{equation}

On modern CNN architectures, convolution layers are often coupled with batch-normalization layers that normalize (recentering and rescaling) the inputs of layers to make training faster and improve stability.

\subsection{Convolution primitives}

We detail the different convolution primitives evaluated in this work. Table~\ref{Primtive_resume} sums up performance features compared to the standard convolution.

\begin{figure}
\centering
\includegraphics[width=0.85\textwidth]{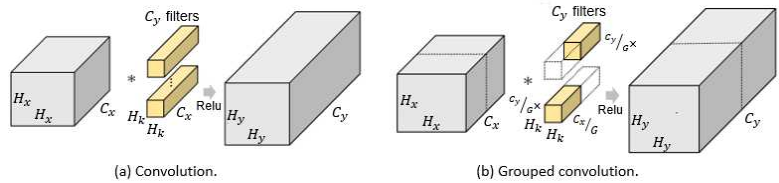}
\caption{From \cite{ioannou2017deep}, standard vs. grouped convolutions: the grouped convolution with 2 groups applies half of the filters to each half of the input channels in order to compute each half of the output channels.} \label{groupedconvolution}
\end{figure}

\begin{table}[t]
\resizebox{\columnwidth}{!}{%
\begin{tabular}{ccccc}

\toprule
Convolution type                           & Parameters                          & Theoretical MACs                                            & Parameters gain                            & Complexity gain                 \\ 
\midrule
Standard                     & $H_k^2\cdot C_x\cdot C_y$           & $H_k^2\cdot C_x\cdot H_y^2\cdot C_y$                  & -                                          & -                               \\ 

Grouped                & $H_k^2\cdot \frac{C_x}{G}\cdot C_y$ & $H_k^2\cdot \frac{C_x}{G}\cdot H_y^2\cdot C_y$        & $\frac{1}{G}$                              & $\frac{1}{G}$                   \\

Depthwise separable  & $C_x\cdot (H_k^2+C_y)$     & $C_x\cdot H_y^2\cdot (H_k^2+C_y)$ & $\frac{1}{C_y}+\frac{1}{H_k^2}$            & $\frac{1}{C_y}+\frac{1}{H_k^2}$ \\

Shift                & $C_x\cdot (2+C_y)$         & $C_x\cdot C_y\cdot H_y^2$                             & $\frac{2}{C_y\cdot H_k^2}+\frac{1}{H_k^2}$ & $\frac{1}{H_k^2}$               \\

Add                  & $H_k^2\cdot C_x\cdot C_y$           & $H_k^2\cdot C_x\cdot H_y^2\cdot C_y$                  & 1                                          & 1                               \\
\bottomrule
\end{tabular}%
}
\caption{Summary of the different primitives. Parameters gain is the ratio between the primitive's number of parameters and the standard convolution. The same applies for theoretical MACs with complexity gain.}
\label{Primtive_resume}
\end{table}
\subsubsection{Grouped convolution}

 was first introduced in the AlexNet paper from Krizhevsky \etal~\cite{krizhevsky2012imagenet} for practical issues, then several works such as Ioannou~\etal~\cite{ioannou2017deep} have studied its effect on the performance of a neural network model. For the standard convolution, all input channels are used to compute an output channel. For a grouped convolution with G groups, each channel of the input and output are associated with a group $G_i$. Then, to compute an output channel of the group $G_i$, only the corresponding input channels are processed, as depicted in Fig.~\ref{groupedconvolution}. Thus, grouped convolutions (also referred as \textit{filter groups}) reduce the number of parameters and MAC operations of the layer by a factor G.

\vspace{-10pt}
\subsubsection{Depthwise separable convolution}
Szegedy~\etal~\cite{szegedy2015going} introduce depthwise separable convolutions with the \textit{Inception} architecture. Depthwise separable convolution replaces the standard convolution by two convolutions: \textit{depthwise} and \textit{pointwise}.
Depthwise convolution is an extreme version of grouped convolution where $G = C_x = C_y$. The problem is that each filter only handles information passed down from one input channel. Pointwise convolution is applied to linearly combine the output channels of the depthwise convolution thanks to $1\times1$ kernels. It also acts as a reduction of the depth of the output tensor $Y$.
\vspace{-10pt}
\subsubsection{Shift convolution}

Even though pointwise convolution is more computationally expensive than depthwise convolution in theory, Jeon~\etal~\cite{jeon2018constructing} notice, with a hardware implementation, that depthwise convolution is more time-consuming than point convolution. They replace depthwise convolution by a shift operation which requires extremely few parameters and less computational power to produce the intermediate feature map $I$:

\begin{equation}
I_{k,l,m}=X_{k+\alpha_m,l+\beta_m,m} \text{  } \\\forall k,l \in [1,H_x],\quad\forall m \in [1,C_x]
\end{equation}
where $\alpha_m$ and $\beta_m$ denote the horizontal and vertical shift assigned to the $m^{th}$ channel of the input feature map.\\
\vspace{-10pt}
\subsubsection{Add convolution}

Multiplication operation consumes, in most cases, more energy than addition operation. Chen~\etal~\cite{chen2020addernet} exploit the fact that convolutions in deep neural networks are cross-correlation measuring the similarity between input and convolution kernel. They propose to replace cross-correlation by $L1$-norm as a similarity measure to perform an \textit{add convolution} as in Eq.\ref{add convolution}.
\begin{equation}\label{add convolution}
Y_{k,l,n}=-\sum_{m=1}^{C_x}\sum_{i=1}^{H_k}\sum_{j=1}^{H_k}\vert W_{i,j,m,n} - X_{k+i-1,l+j-1,m}\vert \quad\forall k,l \in [1,H_y],\quad\forall n \in [1,C_y]
\end{equation}
The output of an add convolution is always negative. Thus, in order to make add convolution compatible with standard activation functions like ReLu, a batch normalization layer following the add convolution layer is needed.

\subsection{Neural network library for Cortex-M MCU}

The challenge of porting neural networks to constrained platforms such as microcontrollers has led to the creation of embedding tools (e.g. TFLM\footnote{\url{https://www.tensorflow.org/lite/microcontrollers}}, N2D2\footnote{\url{https://github.com/CEA-LIST/N2D2}}, STM32Cube MX-AI\footnote{\url{https://www.st.com/en/embedded-software/x-cube-ai.html}} or NNoM\footnote{\url{https://github.com/majianjia/nnom}}). Those tools support standard convolution as well as depthwise separable convolutions layers. TFLM and STM32Cube MX-AI support floating point operations, 16 and 8 bits integer operations while NNoM supports only 8 bits integer operations. Furthermore, for Cortex-M4 and Cortex-M7 MCUs (with Digital Signal Processing extensions), SIMD instructions can be used for the computation of different primitives by integrating the middleware CMSIS-NN~\cite{lai2018cmsis} to those tools. For our study, the open source NNoM library was chosen due to its good performance and its ease of customization.

\section{Implementation}
In this section, we present the implementation details of NNoM and CMSIS-NN convolution on which our implementations of the different primitives are based. Furthermore, we detail the differences of implementation between the standard convolution and the optimized primitives.

\subsection{Quantization}

Quantization is the process of reducing the precision of weights, biases, and activations in order to reduce the memory footprint. NNoM library uses 8 bits quantization for the weights, biases, and activations with a uniform symmetric powers-of-two quantization scheme as in Eq.~\ref{eq_powers_of_two}.
\begin{equation}
   dec = ceil\Big(log_{2}\big(max(|X_{f}|)\big)\Big) \text{  ;  }
   x_{i} = floor\big(x_{f}\cdot 2^{(8-1)-dec}\big)
\label{eq_powers_of_two}
\end{equation}
where $X_{f}$ is a 32 bits floating point tensor, $x_{f}$ a value of $X_f$, $x_{i}$ its 8 bits quantized version and $2^{dec}$ is the scale of quantization. Because this scale is a power of 2, the convolution operation only requires integer addition, multiplication and bit shifting, but no division (see Algorithm \ref{alg:conv_innerloop}, left). This computation process is used for grouped and shift convolutions because of their similarity to standard convolution. We adapt it to add convolutions as presented in Algorithm \ref{alg:conv_innerloop} (right).

\begin{algorithm}
\caption{Inner loop of convolution (left) and add convolution (right) without bias}
\label{alg:conv_innerloop}
\textbf{Input :} individual weight w, power-of-2 scale of weight $dec_{weight}$, one input value x, power-of-2 scale of input $dec_{input}$, power-of-2 scale of output $dec_{output}$
\begin{multicols}{2}
    \begin{algorithmic}[1]
    \STATE $output \gets i\cdot w$
    \STATE $shift_{output} \gets dec_{weight} + dec_{input} - dec_{output}$
    \STATE $output \gets output >> shift_{output}$
    \STATE Return output
    \end{algorithmic}
    \columnbreak
    \begin{algorithmic}[1]
    \STATE $shift \gets |dec_{input}-dec_{weight}|$
    \IF{$dec_{input}>dec_{weight}$}
        \STATE $output \gets -|i-(w<<shift)|$
        \STATE $shift_{output} \gets dec_{input} - dec_{output}$
    \ELSIF{$dec_{input}<dec_{weight}$}
        \STATE $output \gets -|(i<<shift)-w|$
        \STATE $shift_{output} \gets dec_{weight} - dec_{output}$
    \ELSE
        \STATE $output \gets -|i-w|$
        \STATE $shift_{output} \gets dec_{weight} - dec_{output}$
    \ENDIF
    \STATE $output \gets output >> shift_{output}$
    \STATE Return output
        \end{algorithmic}
\end{multicols}
\end{algorithm}
\vspace{-20pt}

\subsection{Batch normalization folding}

For convolutions, NNoM library uses the batch normalization folding proposed by Jacob~\etal~\cite{jacob2018quantization}. By merging convolution layers and batch normalization layers, this method accelerates the inference without accuracy drop. Batch normalization folding can be applied for the computation of grouped and shift convolutions but is not suitable fot add convolution.

\subsection{Im2col algorithm with SIMD instructions}

In order to accelerate convolutions, the CMSIS-NN middleware~\cite{lai2018cmsis} use the image to column (im2col) algorithm~\cite{chellapilla2006high}. A first step is to sample patches from the input, flatten and stack them as columns of a matrix $M$. Each filters of the convolution weight $W$ are also flattened and stacked as rows of a matrix $N$. In the second step, the output is computed with the matrix multiplication $Y=M.N$.
\\
To deal with the increased memory footprint of im2col, Lai~\etal~\cite{lai2018cmsis} limit the number of patches processed at the same time to 2. The matrix multiplication is computed using 2 filters simultaneously to maximize the data reuse at the register file level on ARM Cortex-M. Furthermore, Lai~\etal~\cite{lai2018cmsis} use the parallelized multiply-accumulate instruction \texttt{\_\_SMLAD} to speed up the matrix multiplication.

For grouped convolution, we apply Lai~\etal~\cite{lai2018cmsis} algorithm to each group. For shift convolution, we modify the first step of im2col to sample a patch with different shifts for each input channel. We did not implement a SIMD version of add convolutions because there is no instructions similar to \texttt{\_\_SMLAD} adapted to add convolutions.

\section{Experimental characterisations}
The experiments are carried out on a typical 32-bit MCU platform, the Nucleo STM32F401-RE, based on Cortex-M4 that supports SIMD instructions. Unless specified, the compiler is arm-none-eabi-gcc (version 10.3) with the optimization level sets to \texttt{Os} and the MCU's frequency is fixed at 84 MHz. The software STM32CubeMonitor-Power\footnote{\url{https://www.st.com/en/development-tools/stm32cubemonpwr.html}} is used to measure the electric current of the MCU. We multiply it by the supply voltage (i.e. 3.3 V) and integrate it over the duration of an inference to obtain the inference's energy consumption.

\subsection{Influence of the primitive parameters}
\subsubsection{Protocol}

To evaluate the influence of a parameter (i.e. kernel size, input width...), we consider a layer with every other parameters fixed excepted the concerned one. The experiment plan is defined in table \ref{tab2}. We measure the latency and energy consumption over 50 inferences (average) with randomized inputs. Results are presented in Fig.\ref{benchmark}.

\subsubsection{Results without SIMD instructions}
\begin{figure}
\includegraphics[width=\textwidth]{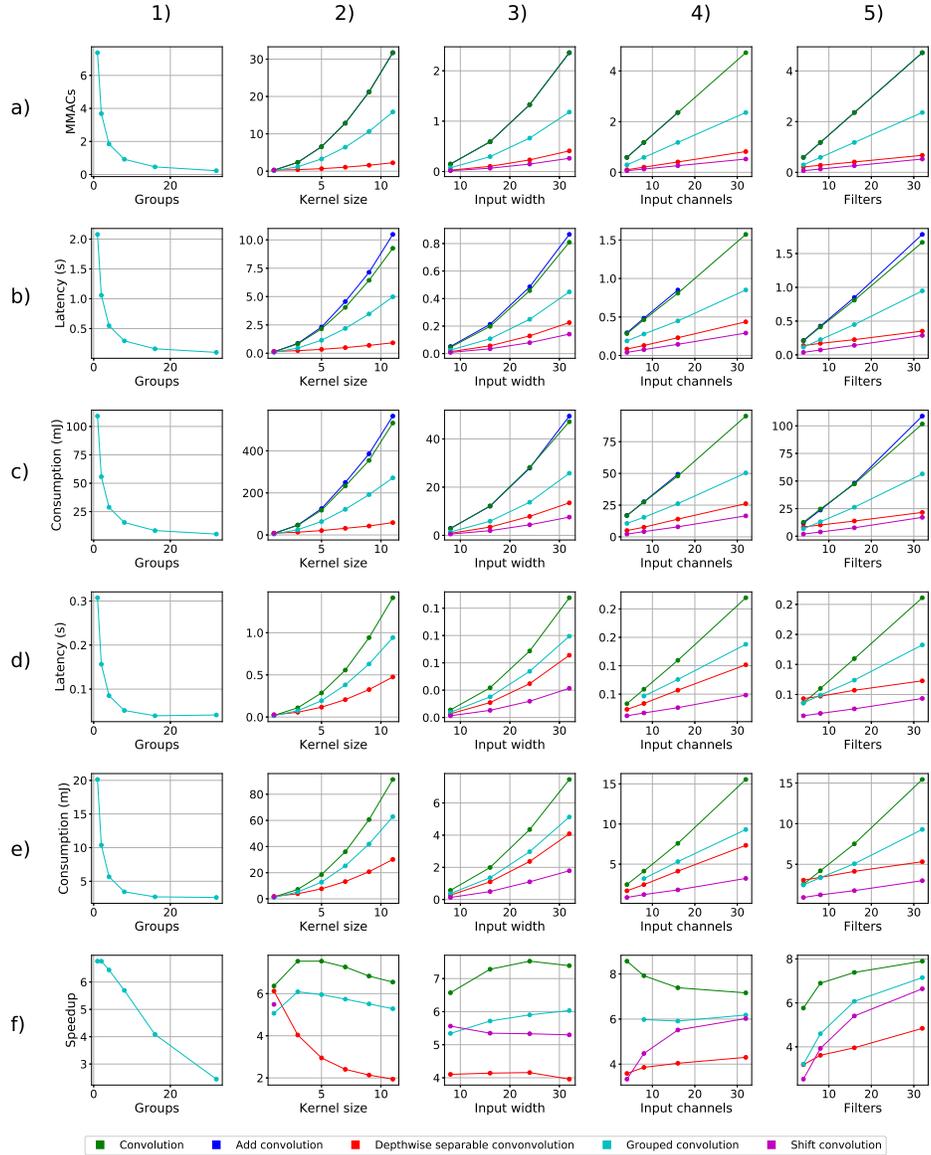}
\vspace{-20pt}
\caption{Influence of the 1) number of groups, 2) kernel size, 3) input width, 4) number of input channels and 5) filters on a) theoretical MACs, b) latency without SIMD instructions, c) energy consumption without SIMD instructions, d) latency with SIMD instructions and e) energy consumption with SIMD instructions and f) speedup for different primitives. The different implementations fit the theory. Using SIMD instructions enables faster and less energy consuming inferences. The speedup of the im2col algorithm varies according to the primitives and their parameters.} \label{benchmark}
\end{figure}

We observe in Fig. \ref{benchmark}.a-c that our implementation fits the theory (Table~\ref{Primtive_resume}). For example, the theoretical MACs, latency and energy consumption increase quadratically with the kernel size (Fig \ref{benchmark}.2.a, Fig \ref{benchmark}.2.b and Fig \ref{benchmark}.2.c). More specifically, there is a linear relationship between the MACs, latency and consumption. A linear regression leads to scores of 0.995 and 0.999 respectively. Add convolutions are slightly less efficient than convolutions despite the same number of MACs. This is explained by the quantization scheme of add convolution and the additional batch normalization layer.

\begin{table}[h!]
\centering
\begin{tabular}{cccccc}
\toprule
Experiment & Groups & Kernel size & Input width & Input channel & Filters \\ 
\midrule
1          & 1-32   & 3           & 10          & 128           & 64      \\ 
2          & 2      & 1-11        & 32          & 16            & 16      \\ 
3          & 2      & 3           & 8-32        & 16            & 16      \\ 
4          & 2      & 3           & 32          & 4-32          & 16      \\ 
5          & 2      & 3           & 32          & 16            & 4-32    \\ \bottomrule
\\
\end{tabular}
\caption{Primitive parameters for the different experiments.}
\end{table}

\subsubsection{Effect of SIMD instructions}

\begin{figure}
\includegraphics[width=\textwidth]{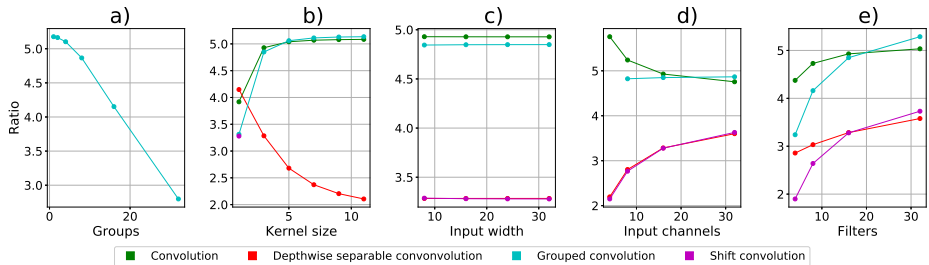}
\vspace{-10pt}
\caption{Influence of the a) number of groups, b) kernel size, c) input width, d) number of input channels and e) filters on the ratio of memory access without SIMD instructions by the memory access with SIMD instructions (normalized by MACs) for different primitives.} \label{ratio}
\end{figure}

Using SIMD instructions decreases the latency (Fig \ref{benchmark}.d) and energy consumption (Fig \ref{benchmark}.e) of the different primitives. Our implementation with SIMD instructions also fits the theory. But latency is more relevant to estimate the layer’s energy consumption (regression score of 0.999) than theoretical MACS (regression score of 0.932). This loss of linearity is related to the varying speedup of the im2col algorithm with respect to the primitives and their parameters (Fig \ref{benchmark}.f). A possible explanation is in the data reuse exploitation by the im2col algorithm. To verify this, we measure the number of memory access in those programs. Fig. \ref{ratio} shows the variation of the ratio of memory access without SIMD instructions by the memory access with SIMD instructions (normalized by MAC) for different parameters and primitives. We observe in Fig. \ref{ratio} the same variations as in Fig. \ref{benchmark}.f . Thus, data reuse contributes strongly to the speed up of algorithms using SIMD instructions. However, convolutions and grouped convolutions have similar ratio in Fig. \ref{ratio} but different speedup in Fig. \ref{benchmark}.f . Other factors such as memory access continuity and padding are to be taken into account to explain the performance of these programs.

\vspace{-12pt}

\subsection{Influence of other factors}
For the following experiments, we fix the number of groups at 2, the kernel size at 3, the input width at 32, the input channel at 3 and the filters at 32.

\begin{figure}[t!]
\includegraphics[width=\textwidth]{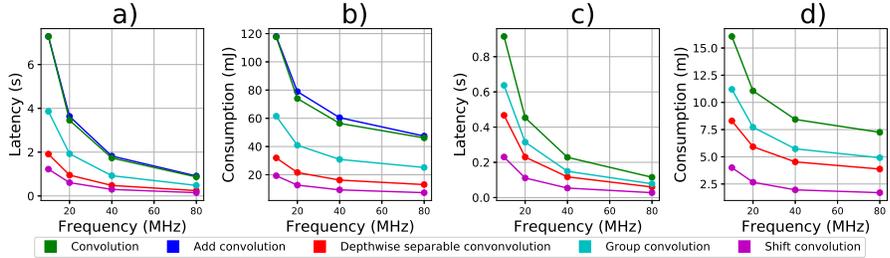}
\vspace{-10pt}
\caption{Influence of the MCU frequency on latency, energy consumption without SIMD instructions (a and b) and with SIMD instructions (c and d).} \label{frequency}
\end{figure}

\subsubsection{Influence of frequency}
We perform inferences on a frequency range from 10 to 80 Mhz (see Figure \ref{frequency}). Latency is inversely proportional to the frequency as expected. Power consumption increases with frequency (see Table \ref{tab2}) but to a lesser degree than the decrease of latency. Thus, using the maximum frequency of ARM Cortex-M MCUs lowers the inference's energy consumption.
\vspace{-10pt}
\begin{table}
\centering
\begin{tabular}{ccccc}
\toprule
        & 10 MHz  & 20 MHZ  & 40 MHz   & 80 MHz   \\ \midrule
No SIMD & 16.16 & 21.59 & 32.83  & 52.09 \\ \midrule
SIMD    & 17.57 & 24.66 & 37.33 & 62.75 \\ \bottomrule
\\
\end{tabular}
\caption{Average power consumption (mW) at different frequencies.}\label{tab2}
\end{table}
\vspace{-40pt}

\subsubsection{Influence of optimization level}
We perform a convolution inference with two different optimization levels (\texttt{O0} and \texttt{Os}). As seen in table \ref{tab3}, the compiler optimization has an important effect on the layer performance. Using \texttt{Os} level accelerates the inference by a factor 1.52. This impact is emphasized with the use of SIMD instructions (factor 9.81). Without optimization, the use of SIMD instructions can even increase the layer's energy consumption as using SIMD instructions increases the average power consumption.
\vspace{-10pt}
\begin{table}[h!]
\resizebox{\textwidth}{!}{%
\begin{tabular}{cccccc}
\toprule
                                               & Optimization level & Latency (s) & Consumption (mJ) & Optimization Speedup & SIMD Speedup \\ \midrule
                                               
\multicolumn{1}{c}{\multirow{2}{*}{No SIMD}} & \texttt{O0}                & 1.26        & 63.9            & -                    & -            \\ 
\multicolumn{1}{c}{}                         & \texttt{Os}                & 0.83        & 45.7             & 1.52                 & -            \\ \midrule
\multicolumn{1}{c}{\multirow{2}{*}{SIMD}}    & \texttt{O0}                & 1.08        & 82.0             & -                    & 1.17         \\ 
\multicolumn{1}{c}{}                         & \texttt{Os}                & 0.11        & 7.2              & 9.81                 & 7.55         \\ \bottomrule
\\
\end{tabular}%
}
\caption{Effect of optimization level on inference performance for convolution.} \label{tab3}
\end{table}
\vspace{-40pt}

\section{Conclusion}
In this paper, we implement and benchmark several state-of-the-art convolution primitives for ARM Cortex-M microcontrollers. Our benchmark shows that for microcontrollers which cannot use SIMD instructions, theoretical MACs is a relevant indicator to estimate the layer energy consumption. For microcontrollers which use SIMD instructions, latency is preferred over theoretical MACS to estimate the layer energy consumption while using SIMD instructions. We explain this by the varying efficiency of the im2col algorithm, from CMSIS-NN, depending on the layers and highlight the role of data reuse in this performance gap. Furthermore, we study the influence of external parameters to the convolution algorithms such as the compiler optimization and the MCU frequency. Our experiments highlight the major impact of the compiler optimization on the layers performance while using SIMD instructions, and show that running the inference at maximum frequency decreases the layer's energy consumption. Our work opens up new possibilities for neural architecture search algorithms.

\subsubsection{Author Contribution} Nguyen, Moëllic and Blayac conceived and planned the study. Nguyen carried out the experiments and performed the analysis. Nguyen and Moëllic wrote the manuscript with inputs from all authors.

\section*{Acknowledgments}
Part of this work was done with the support of ID-Fab (Prototyping platform: project funded by the European Regional Development Fund, the French state and local authorities).
This work benefited from the French Jean Zay supercomputer thanks to the \textit{AI dynamic access} program. This collaborative work is partially supported by the IPCEI on Microelectronics and Nano2022 actions and by the European project InSecTT\footnote{\url{www.insectt.eu}: ECSEL Joint Undertaking (876038). The JU receives support from the European Union’s H2020 program and Au, Sw, Sp, It, Fr, Po, Ir, Fi, Sl, Po, Nl, Tu. The document reflects only the author’s view and the Commission is not responsible for any use that may be made of the information it contains.} and by the French National Research Agency (ANR) in the framework of the \textit{Investissements d’Avenir} program (ANR-10-AIRT-05, irtnanoelec).

%
%
%
%
\bibliographystyle{splncs04}
\bibliography{bib/references.bib}
\end{document}